# Topic Modelling of Empirical Text Corpora:
# Validity, Reliability, and Reproducibility in Comparison to Semantic Maps

Tobias Hecking[a]* and Loet Leydesdorff [b]

**Abstract**
Using the 6,638 case descriptions of societal impact submitted for evaluation in the Research Excellence Framework (REF 2014), we replicate the topic model (Latent Dirichlet Allocation or LDA) made in this context and compare the results with factor-analytic results using a traditional word-document matrix (Principal Component Analysis or PCA). Removing a small fraction of documents from the sample, for example, has on average a much larger impact on LDA than on PCA-based models to the extent that the largest distortion in the case of PCA has less effect than the smallest distortion of LDA-based models. In terms of semantic coherence, however, LDA models outperform PCA-based models. The topic models inform us about the statistical properties of the document sets under study, but the results are statistical and should not be used for a semantic interpretation—for example, in grant selections and micro-decision making, or scholarly work—without follow-up using domain-specific semantic maps.

[a] * **corresponding author** ; University of Duisburg-Essen, Department of Computer Science and Applied Cognitive Science, Lotharstraße 63, 47057 Duisburg, Germany; hecking@collide.info
[b] University of Amsterdam, Amsterdam School of Communication Research (ASCoR), PO Box 15793, 1001 NG Amsterdam, The Netherlands; loet@leydesdorff.net

# 1. Introduction

The capability of topic models to structure large text corpora into latent topics is important for modern information retrieval. Topic models hence are one of the major areas of research in text mining. In recent years, however, topic models are not only used in their original fields of applications (such as document retrieval in the information sciences), but have gained momentum in other domains such as digital humanities, journalism, and policy making. For example, in their study entitled "Quantitative analysis of large amounts of journalistic texts using topic modelling," Jacobi, van Atteveldt, & Welbers (2016, at p. 89) formulate the challenge of topic modelling as follows:

> "The huge collections of news content which have become available through digital technologies both enable and warrant scientific inquiry, challenging journalism scholars to analyse unprecedented amounts of texts. We propose Latent Dirichlet Allocation (LDA) topic modelling as a tool to face this challenge. LDA is a cutting-edge technique for content analysis, designed to automatically organize large archives of documents based on latent topics, measured as patterns of word (co-)occurrence."

Latent Dirichlet Allocation (LDA; Blei, Ng, and Jordan, 2003) inspired the development of a number of similar methods; it belongs to a family of probabilistic models in which a document is modelled as a probability distribution of topics, while each topic is considered as a distribution of words. LDA LDA and similar models are increasingly used in research evaluation; for example, in social-impact studies and by granting organizations.

In this study, we address the questions of the reliability and validity of LDA-based topic models. Can these models carry the science-policy decisions when the latter are legitimated in terms of these models? In the U.S.A., for example, both the National Institute of Health (NIH; Talley *et al*., 2011) and the National Science Foundation (NSF; Nichols, 2014) experimented with "topic models" for organizing their research-grants portfolios. The Research Evaluation Framework (REF 2014) in the UK commissioned a topic model for organizing the social impact statements of the research output under evaluation (Grant et al., 2015).

The main problem with analyzing large corpora of texts that are (almost by definition) beyond the human capacity to comprehend by reading, has remained the validity of the results (Grimmer & Stewart, 2013). Words are so flexible that one can almost always provide an interpretation of a groups of words *ex post*. When words spread across categories, one can, for example, consider these words as "methodological" (Draux & Szomszor, 2017, at p. 12). However, such an explanation remains *ad hoc*.

In the NSF model, for example, thousand topics were constructed on the basis of 170,000 awards granted between 2000 and 2012 (Gretarsson et al., 2012). In an evaluation of this model, Nichols (2014, at p. 747) concluded that 89% of the awards granted by the directorate of the Social and Behavioral Sciences were classified as "interdisciplinary research." Leydesdorff & Nerghes (2017, at p. 1034) raised the question of whether this "interdisciplinarity" were perhaps a consequence of the mixing of disciplinary terminologies by the topic model itself? These authors found statistically significant differences between the outcomes of co-word models and topic models, while using the same input data, lists of words, and stop-words in both models. The results of the co-word model were often interpretable, while the results of the topic models were counter-intuitive.

For example, in a set of documents about research evaluation the words "peer" and "review" which co-occur in "peer review" were attributed to different clusters by the topic model as were "university" and "ranking" despite the topic of "university ranking" prominently visible in the co-word map and an issue under discussion in the document set. For the purpose of topic modeling, however, sets of the order of 1,000 documents are perhaps too small. In this study, we upscale the study of Leydesdorff & Nerghes (2018) to a similar experiment using the above-mentioned 6,638 case descriptions of societal impact of research submitted for evaluation in the context of the Research Excellence Framework (REF 2014; Grant, 2015).

The vision behind topic modelling echoes the program of "co-word mapping" in Science and Technology Studies (STS) by Callon *et al.* (1983; cf. Latour, 1986; Law & Lodge, 1984) and "semantic maps" in artificial intelligence by Landauer, Foltz, & Laham' (1998; cf.van Atteveldt, 2008). However, LDA is provided with user-friendly interfaces which are easily applicable for non-experts (e.g., at https://code.google.com/p/topic-modeling-tool/ since 2013). It is also included in several software tools (e.g., the Stanford Topic Modeling Toolbox at https://nlp.stanford.edu/software/tmt/tmt-0.4/; Ramage, Rosen, Chuang, Manning, & McFarland, 2009), which further contributes to its wide-spread usage. Furthermore, LDA in MALLET[3] is integrated in programs for semantic network analysis such as Diesner *et al*.'s (2015) program ConText for semantic mapping. Thus, the instrument of topic modeling has become intensively used in both scholarly and political contexts, making the validity and reliability of the results of LDA models an urgent topic.

The topic model of 6,638 case descriptions of societal impact of research submitted for evaluation in the context of the Research Excellence Framework (REF 2014) study was commissioned to a combined group of researchers at King's College in London and an expert group at Digital Science—an offspring of the Nature Publishing Group/Macmillan. Both the data and the report (Grant, 2015) are available on the internet. The authors of the report programmatically stated (at p. 85) that, in comparison to other techniques, LDA has the following advantages:
> One of the most important aspects of topic modelling as implemented in LDA is that rather than simply basing topics on word features occurring in documents together, it uses contextual information of word occurrences in documents, and so can capture words with similar meanings that are used interchangeably within similar contexts.
> […]
> LDA is the accepted state-of-the-art in topic modelling and is implemented in many standard toolboxes for machine learning.

Are we able to replicate this study given the publicly available data and the transparency of the methodological decisions? We first invited the authors of the REF study at Digital Science to collaborate in such a replication, but we obtained the following answer (personal communication, 4 October 2017):

> Generating the same topic model that King's produced would be practically impossible. Even if we were able to obtain the original source data, the code for preprocessing, and the topic modelling parameters, the output could still differ depending on the software libraries used and their versions. As you may also be aware, most topic modelling implementations will rely on a random seed that may not be known.
>
> To complicate matters further, the researcher responsible for the analysis […] has now left King's. For this analysis, King's did not use the text that is now made available on the case studies website. Those cleaned versions were not available at the time so they made use of the text that could be automatically extracted from the original PDFs.
>
> Our current messaging around topic modelling is that no single topic model is more correct than another, but one may be more suited to answering a particular question than another. The target number of topics is the main consideration here, and usually needs to be made relevant to the likely use-case, with small numbers giving very broad generalisations and higher numbers giving more detail. If granularity is pushed too high, the topics start to degrade into incoherent nonsense. We tune these parameters on a per-dataset basis depending on diversity and volume of text.

---

[3] MALLET (the "MAchine Learning for LanguagE Toolkit") has been developed by a team at UMass Amherst since 2002 (McCallum, 2002). MALLET is available at http://mallet.cs.umass.edu .

In other words, the results of a topic model are (1) anyhow irreproducible because of the random seed and ongoing updates of the hard and software, and (2) the customer can have a considerable say in the results because parameters have to be tuned to the use-case and its objectives.

In our opinion, one can use Gibbs sampling with a fixed seed for circumventing the first problem of the random feed. Given our concerns about the validity of the resulting topics and these reservations about the reproducibility of a model in different runs, however, we focus on the reliability and stability in a space of *possible* solutions, yet using these same case materials and, as much as possible, similar or comparable techniques.

## 2. Methods

As in the previous study of Leydesdorff & Nerghes (2017), we use factor analysis (principal component analysis [PCA] with Varimax rotation) of the word/document matrix for the comparison. Both LDA and PCA can be used to attribute values to documents as cases and words as variables, and both can be used for grouping words or documents. In the case of PCA, one can use factor loadings of the words and factor scores of the documents, respectively.

LDA is a generative model in the sense that probability distributions over topics and words are learned as the one most likely generated in the observed corpus when the data is sampled. In LDA a document is considered a probability distribution of topics and a topic a distribution over the words. The probability of the participation of words and documents in each topic can thus be estimated.

One can compare the differences between the two classifications of words into topics (LDA) or clusters (PCA) using Cramèr's *V* which offers a summary statistics between zero and one based on the chi-square. Despite this comparability of LDA and PCA in terms of the results, the two techniques are very different. LDA is based on a probabilistic model, whereas PCA is based on matrix algebra. PCA can in principle be done with pencil and paper. However, the number of documents and memory requirement are limiting factors in PCA, while LDA can be used for analyzing very large sets.

In the case of PCA the number of factors to be extracted requires as much a decision as the number of topics in LDA. A topic model has to be "trained" in order to make it fit for the purpose of its applications. However, in the case of PCA several statistical tools are available such as scree plots and the percentage of variance explained to guide this decision, while in the case of LDA these choices have to be made in the practical context of the application.

In summary, we compare LDA and PCA in terms of the following three main issues:

1. *Stability* of the model. Since training a LDA model requires sampling of probability distributions, models of the same corpus can be expected to differ as seeds of the random number generator vary. In addition to this inherently non-deterministic nature of LDA, the sensitivity of topic models to relatively small corpus changes is another issue: do small changes in drawing a sample have large effects? If a model shows a high sensitivity to minor variations in the samples, the corpus size, and the sampling procedure can have an impact on the results and thus may lead to erroneous conclusions and unwarranted interpretations.

    Topic modelling is well-suited for structuring large document corpora, for example, for building document retrieval systems, especially when reading is beyond the human capacity. In these cases the goal is rather to group similar documents according to latent topics in closed corpora. However, when it comes to topic modelling in sampled and open corpora, stability issues become more salient.

2. *Validity* of the assignment. In the social sciences, one often does not have an external ground-truth of the data to validate the topics in terms of their meanings. Furthermore, assessment of validity is domain-specific, and thus, one needs domain expertise (Mimno, Wallach, Talley, Leenders, & McCallum, 2011). However, if an expert would also be able to specify the topics in a corpus of documents, automatic topic modelling would no longer be needed. Validation and interpretation of the discovered topics is closely coupled with the validity of conclusions based on these models. Thus, possibly incorrect or hardly interpretable models can have an unwanted impact when they are used for decision support.

3. *Interpretation of topics*. Topic models rely on representations of words and documents in latent spaces. It is commonly assumed that the latent space is semantically meaningful but these assumptions have to be supported by quantitative evaluations of the coherence and meaningfulness of topics (Chang, Gerrish, Wang, Boyd-Graber, & Blei, 2009).

## 3. Data

We use the 6,638 "impact case studies" of REF-2014 which are available for download at http://impact.ref.ac.uk/CaseStudies/Results.aspx?val=Show%20All . The texts in the column of the spreadsheet headed "Summary of the impact" were transformed into lower case and pre-processed as in the original study (Grant et al., 2015) by using stemming (Porter, 1980), stop-word removal, and removal of punctuation. This leads to 30,934 words which occur 517,211 times in the set. Of these words 898 occur more than 100 times. These words occur 352,205 times (86.1% of the 517,211 occurrences) in total. Eight words were removed for technical reasons.

Our basic word/document matrix (Salton & McGill, 1983) thus contains 890 words as column variables attributed to 6,638 documents as rows. This matrix is factor-analyzed using SPSS (v. 22). We also derive from this matrix a cosine-normalized co-occurrence matrix among the 890 words which will be analyzed and visualized using the implementation of the so-called Louvain-algorithm for community-finding in Pajek (Blondel *et al.*, 2008) and VOSViewer for the visualization (van Eck & Waltman, 2010). The cosine is similar to the Pearson correlation underlying the factor analysis, but without the normalization to the mean. Cosine-normalization scales the numbers of word occurrences between zero and one; the resulting visualization is focused on structural components more than without this normalization.

In order to avoid variation in the topics among runs induced by the non-determinism (that is, the initial seed) of LDA, the random seed for Gibbs sampling was fixed so that runs on the basis of the same corpus yield exactly the same result. Using the 6,638 texts as input, we perform LDA with the following parameters: (1) 40 burns in iterations; (2) 1500 iterations; (3) alpha = 50 / # texts; (4) thinning = 50. These values are akin (if not similar) to the ones used by Grant *et al.* (2015) for generating the original topic model.

## 3. Results

### 3.1. Descriptives

Not surprisingly given the sparsity of the matrix, the scree-plot of the factor analysis is very flat: 361 eigenvalues are larger than 1.0 (the default in SPSS). This can hardly be considered as a reduction of the complexity. However, this is a well-known problem when considering texts as bags of words; words are used flexibly. Citations, for example, are more specific than words by an order of magnitude (Braam, Moed, & van Raan, 1991; Leydesdorff, 1989). Decomposition of the cosine-matrix using the Louvain algorithm suggests six to eight distinct communities with modularity $Q = 0.10$. Visual inspection of the scree-plot makes an eight-factor solution also plausible. However, eight factors explain only 3.05% of the variance in the matrix.

For the orientation of the reader Figure 1 provides a visualization of the eight components using the two-mode matrix of 890 words versus eight clusters of words based on PCA (Vlieger &

Leydesdorff, 2011). The factor designation is ours (Table 1). The eight factors (PCA) are compared in Table 1 with an eight-topic solution of LDA in Table 2. Six topics can be unambiguously mapped to topics suggested by factor analysis (Industrial, Medical, Education, Policy, Culture, and Economy). The remaining two topics cannot or only partially be interpreted. The first topic suggests a topic that is distributed as a layer of "methodological" terms distributed across the texts (Draux & Szomszor, 2017, at p. 12), whereas PCA focuses by definition on specific densities.

**Figure 1:** Eight topics based on principal component analysis of the word-document matrix.

**Table 1:** Eight topics identified by Factor Analysis

| cultural | policy advice | medical | industrial | education | automation | climate | urban |
|---|---|---|---|---|---|---|---|
| **public** | polici | clinic | company | educ | softwar | wa | manag |
| **culture** | debat | patient | Spinout | school | method | million | local |
| **audienc** | law | treatment | technologi | teacher | model | year | plan |
| **engag** | govern | trial | Ltd | learn | data | 2008 | environment |
| **exhibit** | right | disea | company | train | ar | 2011 | climat |
| **museum** | reform | therapi | Product | teach | system | per | region |
| **art** | committee | guidelin | commerciali | children | tool | 2013 | urban |
| **histori** | influenc | drug | commerci | profession | industri | estim | citi |
| **audience** | research | care | Patent | practic | comput | 2012 | sustain |
| **artist** | legisl | Nh | Market | servic | algorithm | 2010 | environ |

**Table 2:** Eight topics identified by LDA

| "textual" | industrial | medical | education | policy | climate | culture | economy |
|---|---|---|---|---|---|---|---|
| **thi** | product | health | Educ | polici | univers | public | manag |
| **work** | new | patient | servic | nation | world | cultur | model |
| **studi** | company | clinic | practic | inform | sinc | understand | improv |
| **intern** | industry | treatment | programm | govern | thi | work | provid |
| **case** | design | care | support | influenc | led | media | assess |
| **signific** | technologi | improv | commun | intern | 2008 | engag | chang |
| **practice** | system | now | local | european | major | project | environment |
| **within** | commerci | Led | profession | chang | base | new | data |
| **contribut** | market | result | school | debat | 8212 | art | tool |
| **relat** | process | drug | social | contribut | million | histori | new |

## 3.2. Stability of topics

In both LDA and PCA models based on the word-document matrix, one can attribute a weight indicating the belonging to a topic or factor to words. It is common practice to consider the top-10 words of each topic in presentations. As noted above, a problem may arise when topics are modelled on the basis of empirically sampled text collections: subsequent samples can be considered as an approximation of the actual topics in the corpus materials under study, although a human reader may not recognize different sets of words as representations of the same topic. This problem, however, is similarly the case for LDA and PCA.

How sensitive are these two type of models for small changes in the sampling? This can be studied by removing texts from the sample using random drawings. We first selected 1000 documents from the REF corpus randomly ($C_{rem}$). From this sample, twenty new text corpora in steps of fifty: $C_{-50}, C_{-100}, C_{-150}, ..., C_{-1000}$ were created by removing the first 50, 100, 150, … 1000 documents of $C_{rem}$ from the original corpus ($C_{orig}$). Topics were then modelled for each of these 20 new corpora and compared to the topics derived from the original text collection $C_{orig}$ in order to investigate the impact of these variations in the text sampling on the outcome.

An LDA-based model *m* differs from a PCA-based clustering because the words are not partitioned, but words can be part of multiple topics. Using LDA, each topic model *m* comprises *T* topics *t=(w1, w2, ...wS)*; each topic is represented by a word vector of length *S*. Using Equation 1, however, it is still possible to formulate an adapted version of the so-called purity measure *topSim* for comparing a topic model $m_{test}$ to a reference model $m_{ref}$ by calculating the overlap of the top *S* words of a topic in $m_{test}$ and the best matching topic in $m_{ref}$. The values of *topSim* range between 0 and 1.

$$topSim(m_{test}, m_{ref}) = \frac{1}{TS} \sum_{t_i \in m_{test}} \text{argmax}_{t_j \in m_{ref}} |t_i \cap t_j|$$
(1)

Note that *topSim* is not a strict similarity function since it is not symmetric, i.e., $topSim(m_{test}, m_{ref}) \neq topSim(m_{ref}, m_{test})$. However, it provides an indication of how much a given model deviates from the reference model. In this study, the reference model is given as the topics derived from the entire text corpus $C_{orig}$, and the models to test are those derived from the twenty reduced samples $C_{-50}, C_{-100}, C_{-150}, ..., C_{-1000}$.

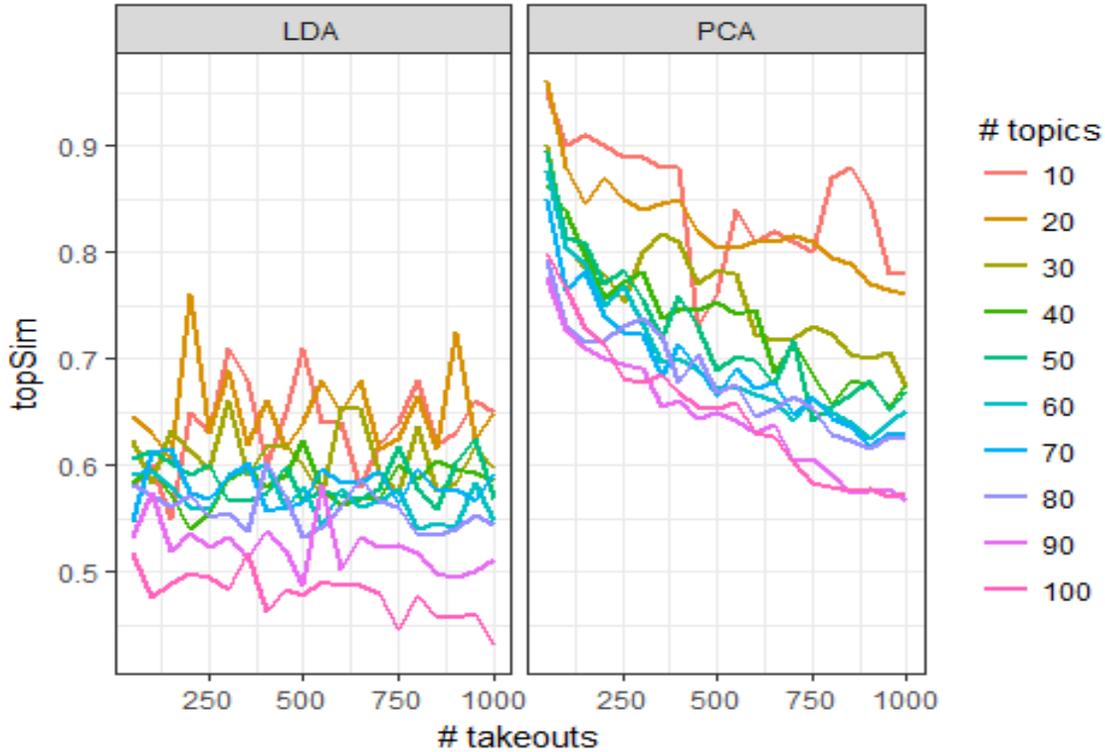

**Figure 2:** Similarity of topic and factor models derived from samples ($C_{-50}, C_{-100}, C_{-150}, ..., C_{-1000}$) compared with the original collection ($C_{orig.}$)

Figure 2 shows the values of $topSim(m_{test}, m_{ref})$ for the twenty reduced text corpora for different numbers of topics generated by LDA (left-hand panel) and PCA of the word-document matrix (right-hand panel), respectively. As can be expected, the similarity between the models tends to decrease the more documents are removed from the original corpus, especially for larger numbers of topics. This tendency is much more salient for PCA-based models than for LDA. However, the deviation of the LDA models based on samples from the reference model is already large when only a small fraction of documents is removed from the original corpus. In the case of LDA, removing only 50 documents can lead to topics very different from the original model, whereas the PCA-based model is more robust and less sensitive for this first intervention.

Furthermore, the more topics are declared the larger the deviation of the topics found in the samples compared with the reference. This result can be expected for the following reason: when more topics are extracted, more degrees of freedom are introduced and therefore topics can be more differentiated. However, the relationship between the number of topics and the sensitivity to corpus changes raised questions when applying topic modelling to the REF study and, therefore, we take this problem one step further : How rapidly does each model decay when the number of topics increases?

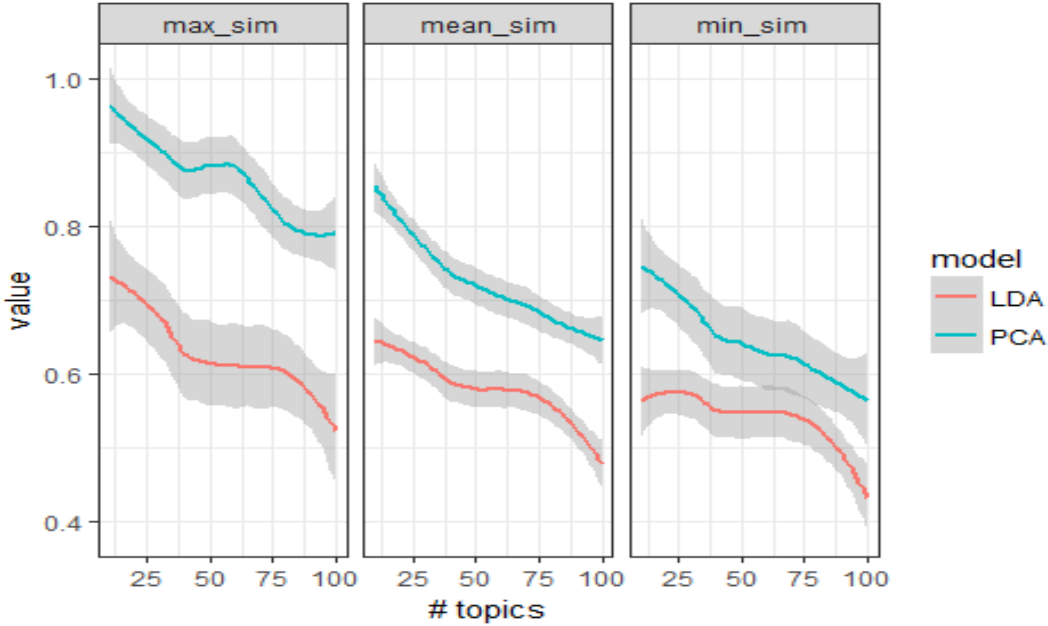

**Figure 3:** Maximum, mean, and minimum topSim across the reduced samples for different number of topics based on LDA and PCA, respectively.

Figure 3 shows the impact of the number of topics on the maximum, mean, and minimum similarity of the different models compared to the reference models for LDA and PCA, respectively. In both cases the decline in the similarity with increasing numbers of topics provides further evidence that the sensitivity to corpus variations is correlated with the number of topics. However, the comparison of LDA and PCA models shows that the *minimum* similarity of a PCA-based model is in most cases still larger than the corresponding *maximum* value for LDA. In other words: the largest distortion by sampling in the case of PCA is smaller than the smallest distortion in the case of LDA. PCA thus outperforms LDA in this respect.

The practical implication of these findings, especially for probabilistic topic models, is that one has to be very careful when applying topic modelling on empirical data sampled as a subset of the corpus in a domain. A small bias in the sampling or incidentally missed documents can have a large impact on the inference of topics and thus on the conclusions and decisions based on the models. These models may not be sufficiently robust for serving as legitimation of decisions.

### 3.3. *Interpretablity and Semantic Coherence*

The background problem of working with empirically sampled data is the absence of ground-truth data that enable us to assess different models using external validity criteria. Thus, internal validity criteria are needed that allow an analyst to obtain insights into how adequate the actual topics of a domain are covered by a topic model. One of these measures is the semantic coherence of topics.

$$C(t; V_t) = \sum_{w_i, w_j \in V_t} \log\left(\frac{D(w_i, w_j) + 1}{D(w_j)}\right) \qquad (2)$$

According to Mimno *et al.* (2011) semantic coherence of a given topic *t* can be measured using Equation 2: the parameter $V_t$ denotes the set of words representing the topic (here the top 10 most associated words); $D(w_i, w_j)$ is the number of documents containing the words $w_i$ and $w_j$; and $D(w_j)$ is the number of documents containing $w_j$. Stevens, Kegelmeyer, Andrzejewski, & Buttler (2012) provide further evidence that this measure is adequate to compare the outcome of different topic modelling approaches.

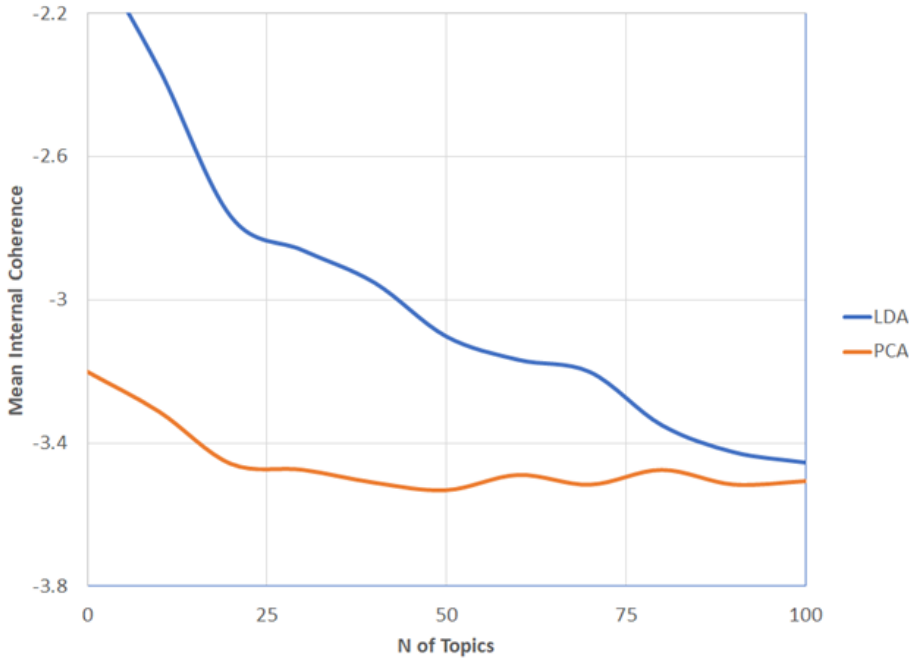

**Figure 4:** Average internal coherence of topics using LDA and PCA, respectively.

The comparison of the average internal coherence of the topic models produced by LDA and PCA for a different number of topics (on the *x*-axis) is shown in Figure 4. The coherence is higher in the case of LDA-based topic models when compared with PCA-based models. However, the internal coherence (IC) of LDA-based models decreases with increase in the number of topics, while in PCA-based models the number of topics does not affect internal coherence.

Stevens *et al.* (2012) found also a high IC for LDA-based models. The decline in the curve for LDA suggests that there is a trade-off between coherently capturing the topics and the sensitivity to the number of topics. However, when we ran a number of tests with eight topics as in Table 2, the topics varied in terms of the top-10 words, but the classification was always significantly the same (with Cramèr's *V* by the order of 0.7, $p < .001$).

## 4. Discussion and Conclusions

The enthusiasm for topic modelling as a technique to summarize large amounts of documents in the format of a limited number of words is not always justified because of the validity problems which are inherent to this methodology. The probabilistic character of the results may easily lead to misunderstandings outside the context of the production of these models. Producers sometimes go along with clients to give the results an interpretation and to "train" the model for choosing parameters leading to results which are plausible. From the perspective of careful decision-making, however, these models can be considered as quicksand on which one should not build. They provide an equivalent to catalogue systems but seem more objective because the generation is computer-assisted.

We raised this question initially in relation to co-word modelling of samples of a size which allowed us to have substantive understanding of the results given our own background (Leydesdorff & Nerghes, 2017). As noted, the results were sometimes very counter-intuitive. Proponents of topic modelling assured us that the problems were generated because of the relatively small sizes of the samples. However, sample sizes should be so large (above 1,000 documents) that human validation is impossible in practice. Added to this is the irreproducibility of topic models and the dependency of a host of parameters which are usually not under control such as new versions of computer programs.

In this study, we controlled for the reproducibility issue. We found that the topic structure is reliably reproduced from run to run, although the lists of top-10 words made visible as indicative may vary. This has increased our trust in the reliability of the technique: the system seems not to get stuck in sub-optima. The validity, however, remains a serious matter of concern. We showed that LDA-based topic models are more sensitive than PCA-based models when relatively small changes are made in the corpus or the number of topics to be extracted. The most drastic distortion of the PCA model had less effect on the results than the most modest distortion of the LDA model. LDA, however, scored much better than PCA on internal cohesion.

In summary, LDA-based models provide reliable statistical results about the corpus under study, which is very appropriate and desirable for structuring more or less "closed" collections of documents in information retrieval. However, if applied to "open-ended" corpora where only samples can be obtained as in the REF study, LDA based topic models are difficult to validate while they may not be valid because of the sensitivity to small variations in the document corpus (Section 3.2 above). Consequently they should not be used as the basis for decision making or intellectual delineations of domains in scholarly works. Models based on co-occurrences of words in word/document matrices may be a preferable alternative in situations where the content itself counts such as in micro-decision making and in scholarly work.

Since this work has shown different strengths and weaknesses of the two modelling approaches, a further question can be how to balance the strengths and weakeneses of probabilistic and coword-based methods. One solution may be, for example, the triangulation of various methods as proposed by T-Lab (Cortini & Tria, 2014). Triangulation of models may an approach to come to more reliable decisions based on unknown or dynamic corpora.

**Limitations of the study**
The results presented in this study give some general insights into the basic properties of topic modelling methods in open corpora when it comes to the question of reliability and interpretability of the results. Stability issues of topic models have been addressed in relation to the number of topics (Greene et al., 2014), and variation in the input vocabulary (De Waal, Barnard, 2008). Agrawal et al. (2014) lists several studies that mention different aspects of topic instability. This study extended these works by investigating the effect of corpora variation on the extraction of topics. However, to come to a general conclusion and guidelines for how to support policy-making by topic models, more extensive experiments with different corpora in more areas of application are required.

In terms of interpretability and semantic coherence only internal validation methods could be used due to a lack of ground-truth topics. It is very difficult to produce such ground-truth from large corpora and a lot of expert knowledge is required. However, the reason for applying topic modelling in science-policy making is that the thematic organisation of the impact of science over several years is not known. If a human being would be able to specify the ground-truth topics, automated analysis of science impact studies would not be needed. A way out of this dilemma could be to combine internal and external validation, where initially detected topics—using topic modelling exploratively—are further empirically validated in follow-up studies using semantic maps.

**References**

Agrawal, A., Fu, W., & Menzies, T. (2018). What is wrong with topic modeling? And how to fix it using search-based software engineering. *Information and Software Technology*, *98*, 74-88.
Blei, D. M., Ng, A. Y., & Jordan, M. I. (2003). Latent dirichlet allocation. *J.Mach.Learn.Res., 3*, 993-1022.
Braam, R. R., Moed, H. F., & van Raan, A. F. J. (1991). Mapping of science by combined co-citation and word analysis. I. Structural aspects. *Journal of the American Society for Information Science, 42*(4), 233-251.
Callon, M. (1986). The sociology of an actor network: the case of the electric vehicle. In M. Callon, J. Law & A. Rip (Eds.), *Mapping the dynamics of science and technology* (pp. 19-34). London: Macmillan.
Callon, M., & Courtial, J.-P. (1989). *Co-Word Analysis: A Tool for the Evaluation of Public Research Policy* Paris: Ecole Nationale Sup,rieure des Mines).
Callon, M., Courtial, J.-P., Turner, W. A., & Bauin, S. (1983). From Translations to Problematic Networks: An Introduction to Co-word Analysis. *Social Science Information 22*(2), 191-235.
Chang, J., Gerrish, S., Wang, C., Boyd-Graber, J. L., & Blei, D. M. (2009). Reading tea leaves: How humans interpret topic models. In *Advances in neural information processing systems*. (pp. 288-296).
Cortini M., Tria S. (2014), Triangulating Qualitative and Quantitative Approaches for the Analysis of Textual Materials: An Introduction to T-Lab, Social Science Computer Review, Vol. 32 no. 4, August 2014, pp. 561-568
De Waal, A., Barnard, E. (2008) Evaluating topic models with stability. In: 19th Annual Symposium of the Pattern Recognition Association of South Africa; retrieved from http://hdl.handle.net/10204/3016 (24 May 2018).
Draux, H., & Szomszor, M. (2017) Topic Modelling of Research in the Arts and Humanities: An analysis of AHRC grant proposals. (pp. https://figshare.com/articles/Topic_Modelling_of_Research_in_the_Arts_and_Humanities/5621260/5621261). London: Digital Research Reports.

Eco, U. (1976). *A Theory of Semiotics (Transl. W. Weaver)*. Bloomington: Indiana University Press.
Grant, J. (2015). *The nature, scale and beneficiaries of research impact: An initial analysis of Research Excellence Framework (REF) 2014 impact case studies*. London: King's College and Digital Science
Greene D., O'Callaghan D., Cunningham P. (2014) How Many Topics? Stability Analysis for Topic Models. In: Calders T., Esposito F., Hüllermeier E., Meo R. (eds) Machine Learning and Knowledge Discovery in Databases. ECML PKDD 2014. Lecture Notes in Computer Science, vol 8724. Springer, Berlin, Heidelberg.
Greimas, A. J. (1983). *Du sens II: Essais sémiotiques*. Paris: Éditions du Seuil.
Gretarsson, B., O'donovan, J., Bostandjiev, S., Höllerer, T., Asuncion, A., Newman, D., & Smyth, P. (2012). Topicnets: Visual analysis of large text corpora with topic modeling. *ACM Transactions on Intelligent Systems and Technology (TIST), 3*(2), 23.
Jacobi, C., van Atteveldt, W., & Welbers, K. (2016). Quantitative analysis of large amounts of journalistic texts using topic modelling. *Digital Journalism, 4*(1), 89-106.
Landauer, T. K., Foltz, P. W., & Laham, D. (1998). An introduction to latent semantic analysis. *Discourse processes, 25*(2), 259-284.
Landauer, T. K., Foltz, P. W., & Laham, D. (1998). An introduction to latent semantic analysis. *Discourse processes, 25*(2), 259-284.
Latour, B. (1986). Visualisation and Cognition: Drawing things together. *Knowledge and Society: Studies in the Sociology of Culture Past and Present, 6*, 1-40.
Latour, B. (1996). On interobjectivity. *Mind, Culture and Activity, 3*(4), 228-245.
Leydesdorff, L. (1989). Words and Co-Words as Indicators of Intellectual Organization. *Research Policy, 18*(4), 209-223.
Leydesdorff, L., & Nerghes, A. (2017). Co-word maps and topic modeling: A comparison using small and medium-sized corpora (N < 1,000). *Journal of the Association for Information Science and Technology, 68*(4), 1024-1035. doi: 10.1002/asi.23740
Mimno, D., Wallach, H. M., Talley, E., Leenders, M., & McCallum, A. (2011). Optimizing semantic coherence in topic models. In *Proceedings of the conference on empirical methods in natural language processing*. (pp. 262-272). Association for Computational Linguistics.
Porter, M. F. (1980). An algorithm for suffix stripping. *Program, 14*(3), 130-137.
Ramage, D., Rosen, E., Chuang, J., Manning, C., & McFarland, D. (2009). *Topic Modeling for the Social Sciences*.
Salton, G., & McGill, M. J. (1983). *Introduction to Modern Information Retrieval*. Auckland, etc.: McGraw-Hill.
Stevens, K., Kegelmeyer, P., Andrzejewski, D., & Buttler, D. (2012). Exploring topic coherence over many models and many topics. In *Proceedings of the 2012 Joint Conference on Empirical Methods in Natural Language Processing and Computational Natural Language Learning*. (pp. 952-961). Association for Computational Linguistics.
Talley, E. M., Newman, D., Mimno, D., Herr II, B. W., Wallach, H. M., Burns, G. A., . . . McCallum, A. (2011). Database of NIH grants using machine-learned categories and graphical clustering. *Nature Methods, 8*(6), 443-444.
van Atteveldt, W. H. (2008). *Semantic Network Analysis: Techniques for Extracting, Representing, and Querying Media Content*. Charleston, SC: BookSurge.

van Eck, N., Waltman, L., van Raan, A., Klautz, R., & Peul, W. (2013). Citation analysis may severely underestimate the impact of clinical research as compared to basic research. *PLoS ONE, 8*(4), e62395-e62395.

Vlieger, E., & Leydesdorff, L. (2011). Content Analysis and the Measurement of Meaning: The Visualization of Frames in Collections of Messages. *The Public Journal of Semiotics, 3*(1), 28-50.

Wilsdon, J., Allen, L., Belfiore, E., Campbell, P., Curry, S., Hill, S., . . . Wouters, P. (2015). The metric tide: report of the independent review of the role of metrics in research assessment and management. 2015. *Publisher Full Text*.